\documentclass[sigconf,natbib=true]{acmart}
\AtBeginDocument{%
  }

\setcopyright{none}
\copyrightyear{2018}
\acmYear{2018}
\acmDOI{XXXXXXX.XXXXXXX}
\acmConference[Conference acronym 'XX]{Make sure to enter the correct
  conference title from your rights confirmation email}{June 03--05,
  2018}{Woodstock, NY}
\acmISBN{978-1-4503-XXXX-X/2018/06}

\usepackage{multirow}
\usepackage[most]{tcolorbox}

\usepackage{booktabs}
\usepackage{colortbl}
\usepackage{xcolor}
\newcommand{\modelname}{Qwen2.5-7B-Instruct}

\newtcblisting{scomtracebox}{
  enhanced,
  breakable,
  colback=blue!3,
  colframe=blue!75!black,
  colbacktitle=blue!75!black,
  coltitle=white,
  title=\textbf{SCoM Trace Example},
  boxrule=0.6pt,
  arc=2pt,
  left=5pt,right=5pt,top=5pt,bottom=5pt,
  listing only,
  listing options={
    basicstyle=\ttfamily\scriptsize,
    breaklines=true,
    breakatwhitespace=false,
    columns=fullflexible,
    keepspaces=true,
    showstringspaces=false
  }
}
\usepackage{placeins}

\begin{document}
\raggedbottom

\title{VIRAASAT: Traversing Novel Paths for Indian Cultural Reasoning}

\author{Harshul Raj Surana}
\affiliation{%
  \institution{AI Institute, University of South Carolina}
  \country{USA}
}

\author{Arijit Maji}
\affiliation{%
  \institution{Indian Institute of Technology Patna}
  \country{India}
}

\author{Aryan Vats}
\affiliation{%
  \institution{Indian Institute of Technology Patna}
  \country{India}
}

\author{Akash Ghosh}
\affiliation{%
  \institution{Indian Institute of Technology Patna}
  \country{India}
}

\author{Sriparna Saha}
\affiliation{%
  \institution{Indian Institute of Technology Patna}
  \country{India}
}

\author{Amit Sheth}
\affiliation{%
  \institution{AI Institute, University of South Carolina}
  \country{USA}
}


\renewcommand{\shortauthors}{Surana et~al.}

\begin{abstract}
 Large Language Models (LLMs) have made significant progress in reasoning tasks across various domains such as mathematics and coding. However, their performance deteriorates in tasks requiring rich socio-cultural knowledge and diverse local contexts, particularly those involving Indian Culture. Existing Cultural benchmarks are 
(i) Manually crafted, (ii) contain single-hop questions testing factual recall, and (iii) prohibitively costly to scale, leaving this deficiency largely unmeasured. To address this, we introduce \textbf{\textit{VIRAASAT}}, a novel, semi-automated multi-hop approach for generating cultural specific multi-hop  Question-Answering dataset for Indian culture. \textbf{\textit{VIRAASAT}} leverages a Knowledge Graph comprising more than 700 expert-curated cultural artifacts, covering 13 key attributes of Indian culture (history, festivals, etc). \textbf{\textit{VIRAASAT}} spans all 28 states and 8 Union Territories, yielding more than 3,200 multi-hop questions that necessitate chained cultural reasoning.
We evaluate current State-of-the-Art (SOTA) LLMs on \textbf{\textit{VIRAASAT}} and identify key limitations in reasoning  wherein fine-tuning on Chain-of-Thought(CoT) traces fails to ground and synthesize low-probability facts. To bridge this gap, we propose  a novel framework named  Symbolic Chain-of-Manipulation (SCoM). Adapting the Chain-of-Manipulation paradigm, we train the model to simulate atomic Knowledge Graph manipulations internally. SCoM teaches the model to reliably traverse the topological structure of the graph. Experiments on Supervised Fine-Tuning (SFT) demonstrate that SCoM outperforms standard CoT baselines by up to 20 \%.
We release the \textbf{\textit{VIRAASAT}} dataset along with our findings, laying a strong foundation towards building Culturally Aware Reasoning Models.
\end{abstract}

\ccsdesc[500]{Computing methodologies~Natural language processing}
\ccsdesc[300]{Computing methodologies~Information extraction}
\ccsdesc[300]{Computing methodologies~Language resources}
\keywords{NLP, Dataset, Culture, Question Answering, Reasoning, Benchmarking}
\maketitle

\begin{figure*}[t]
\centering
\includegraphics[width=0.95\textwidth]{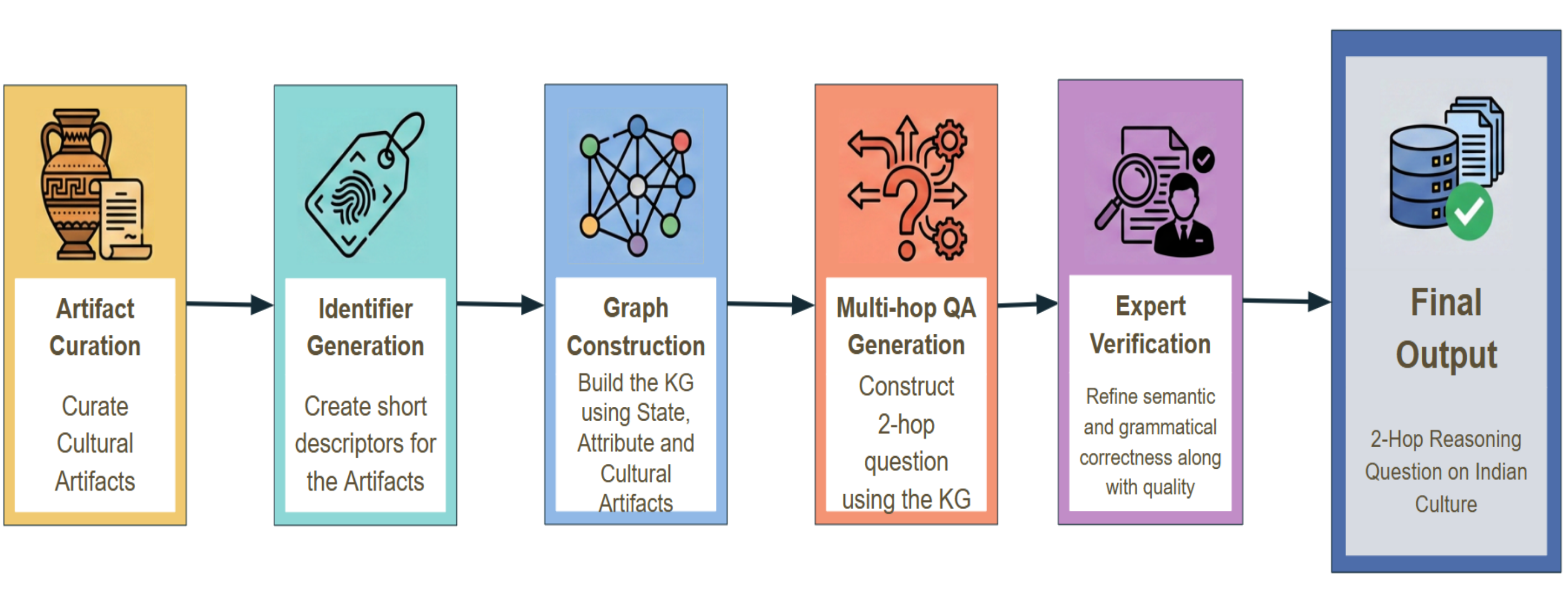}

\caption{\textbf{\textit{VIRAASAT}} dataset construction pipeline: artifact curation and identifier creation, Knowledge Graph construction, template-based 2-hop question generation, and expert verification for semantic and grammatical quality.}
\label{fig:viraasat-dataset-pipeline}
\end{figure*}

\section{Introduction}
Large Language Models (LLMs) have emerged as excellent tools in modern computing, leading to significant advancements across a wide spectrum of Natural Language Processing (NLP) tasks such as Question Answering, Information Retrieval, and content generation \cite{zhao2025surveylargelanguagemodels, minaee2025largelanguagemodelssurvey, kaddour2023challengesapplicationslargelanguage, brown2020languagemodelsfewshotlearners}. Industries are constantly integrating these technologies across various large-scale products and multi-layered business workflows, enabling them to grow to a wider scale with faster communications along with improved decision-making \cite{10822885}. These developments raise a crucial need for LLMs to be sensitive and aware of the vast socio-cultural and linguistic diversities of the different populations consuming and interacting with this technology \cite{plum2025identityawarelargelanguagemodels, adilazuarda-etal-2024-towards, 10.1007/978-3-031-60875-9_18}. However, when tasked with domains requiring rich, socio-cultural knowledge and nuanced local contexts, their capabilities deteriorate significantly \cite{AKSOY2025100172, 10.1162/COLI.a.14, liu-etal-2024-multilingual}. This disparity is particularly acute in the context of Indian culture, a domain characterized by immense linguistic and socio-cultural diversity across 28 states encompassing complex, interlinked traditions, with models frequently struggling to reason accurately about the specific rituals, festivals, or historical artifacts inherent to a specific Indian region \cite{maji-etal-2025-sanskriti, maji-etal-2025-drishtikon}.

This failure is largely attributed to the tail-end distribution of cultural entities within pre-training corpora\cite{nyandwi-etal-2025-grounding}. Unlike global concepts, which appear frequently, specific cultural nuances reside in the "long tail" of the data distribution, making them difficult for models to recall and connect accurately. Progress in addressing this deficiency is currently hindered by the limitations of existing evaluation benchmarks. Datasets such as SANSKRITI \cite{maji-etal-2025-sanskriti} and DOSA \cite{seth-etal-2024-dosa} are primarily manually crafted, restricting scalability and diversity. Furthermore, they focus predominantly on single-hop factual recall (e.g., "Where is the Taj Mahal located?"), evaluating a model's memory rather than its reasoning capabilities. Consequently, there is a lack of rigorous benchmarks that evaluate multi-hop reasoning, the ability to bridge multiple distinct entities to derive the answer, necessitating the need to measure and improve cultural intelligence in AI. Moreover, constructing such benchmarks from scratch is highly time-consuming and requires substantial manual effort, often involving extensive cross-disciplinary collaboration.

\paragraph{\textbf{Research Motivation:}} To address this critical gap, we introduce \textbf{\textit{VIRAASAT}}, a novel, semi-automated, symbolically-grounded Question-Answering benchmark dataset specifically designed to evaluate multi-hop reasoning within the Indian cultural context. Unlike previous crowd-sourced works,  \textbf{\textit{VIRAASAT}} is algorithmically constructed from a high-fidelity Knowledge Graph (KG) backbone comprising over 700 expert-curated cultural artifacts. These artifacts span thirteen distinct attributes of Indian culture, namely history, tourism, cuisine, costume,
language, art, festivals, religion, medicine, transport, cities, sports and personalities, thus providing a multi-faceted representation of India’s cultural tapestry across all 28 states and 8 Union Territories. We generate more than 3,200 complex multi-hop questions (e.g., "Which folk dance is performed in the state associated with the Kangra School of Painting?"). The dataset is verified by relevant experts in the cultural domain to ensure correctness, coherence, and quality. This methodology ensures that every question is grounded in a verifiable logical path, shifting the evaluation focus from simple memorization to structured knowledge traversal. 

We evaluate current State-of-the-Art (SOTA) LLMs, Indic LMs (ILMs), and Small LMs (SLMs) across Zero-Shot and Supervised Fine-Tuning (SFT) settings. Our experiments reveal critical limitations across these settings that impair structured and accurate reasoning. Despite fine-tuning, models struggle to retrieve relevant facts and synthesize them into coherent logical chains, ostensibly due to the artifacts belonging to the tail-end distribution. 

To overcome these limitations, we propose \textbf{Symbolic Chain-of-Manipulation (SCoM)}, a Neuro-Symbolic enforced agentic framework. We draw inspiration from the Chain-of-Manipulation framework \cite{qi2025cogcomvisuallanguagemodel}, where the authors decompose a given VQA task into a series of well-defined atomic manipulations (eg. Zoom, Crop, OCR) to facilitate structured decompositional reasoning. We adapt this concept to the symbolic domain of Knowledge Graphs. In SCoM, we generate long-form reasoning traces by training the model to simulate the role of a Knowledge Graph Agent . The model learns to execute a sequence of Atomic Manipulations, discrete cognitive operations that anchor reasoning in the symbolic structure of the graph. These operations include: 
\begin{itemize}
\item \textbf{Entity Grounding:} The explicit disambiguation of a query term to a specific node.

\item \textbf{Relational Retrieval:} The systematic constraining of search space based on graph topology.

\item \textbf{Semantic Resolution:} Validation of the identity of an entity by matching its attributes with the query description.
\end{itemize}

We employ a Student-Teacher framework to inculcate this behavior. A Symbolic Verifier guides the training process and enforces that every reasoning step generated by the model corresponds to a valid traversal of the underlying graph, ensuring more robust and faithful path-based reasoning. The idea is not to spoon-feed the exact facts to the model, but to empower the model to systematically search and retrieve them.

Our contributions in this work are as follows:

\begin{enumerate}
\item \textbf{VIRAASAT Dataset:} We release the first semi-automated, multi-hop QA dataset for Indian culture, grounded in a domain-specific Knowledge Graph to ensure structure and correctness at scale.

\item \textbf{Systematic Benchmarking:} We provide a comprehensive evaluation of current SOTA LLMs, Indic LMs, and Small LMs in Zero-Shot and SFT settings to establish a performance baseline for multi-hop cultural reasoning.


\item \textbf{SCoM:} We propose Symbolic 
Chain-of-Manipulation, a novel reasoning trace generation paradigm that replaces Chain-of-Thought \cite{10.5555/3600270.3602070} with sequential, graph-grounded manipulations. Our experiments demonstrate that performing SFT on SCoM data improves downstream performance by up to 20\% over standard CoT baselines.
\end{enumerate}

\begin{figure*}[t]
  \centering
  \includegraphics[width=0.49\textwidth,height=0.3\textheight,keepaspectratio]{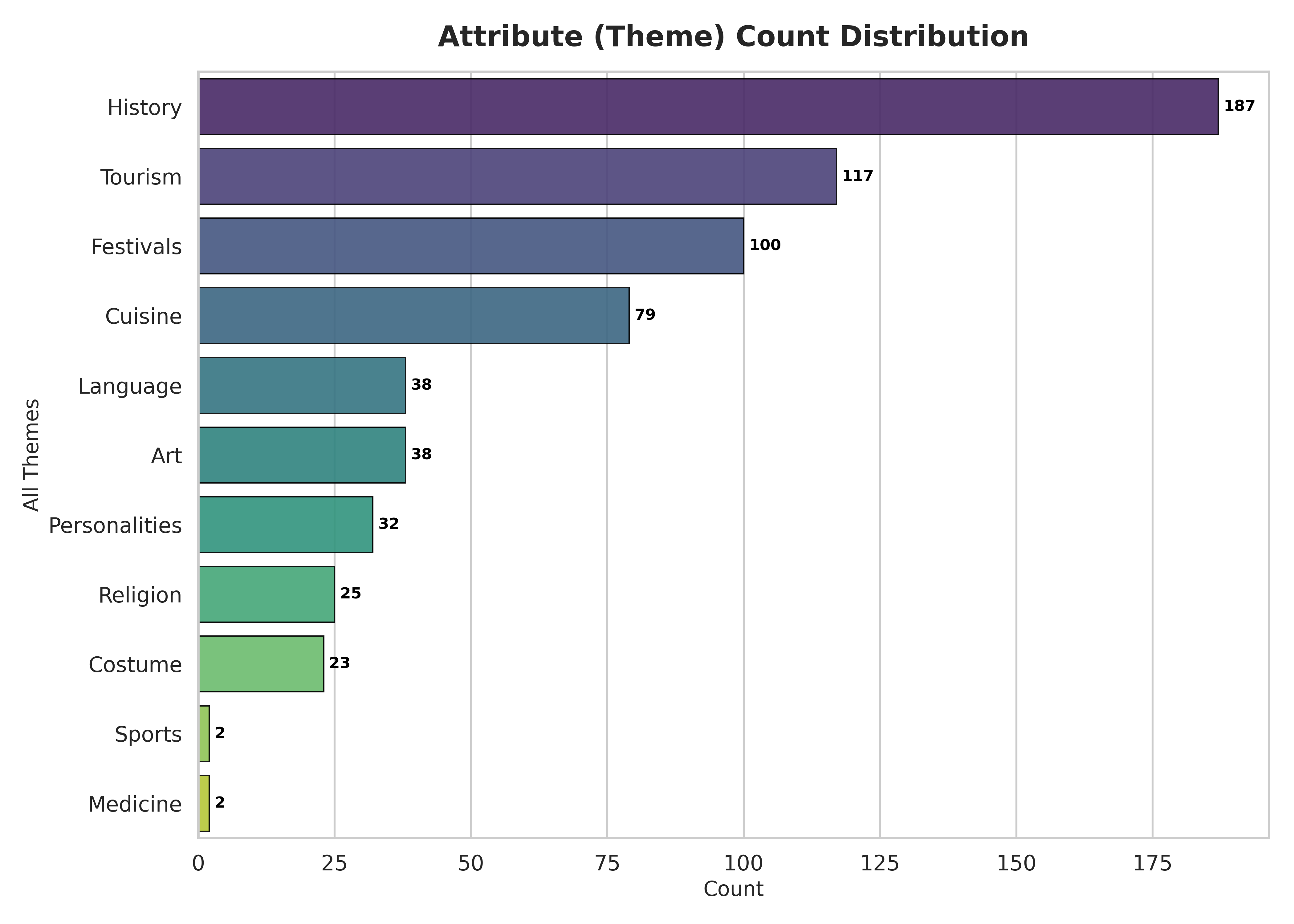}\hfill
  \includegraphics[width=0.49\textwidth,height=0.3\textheight,keepaspectratio]{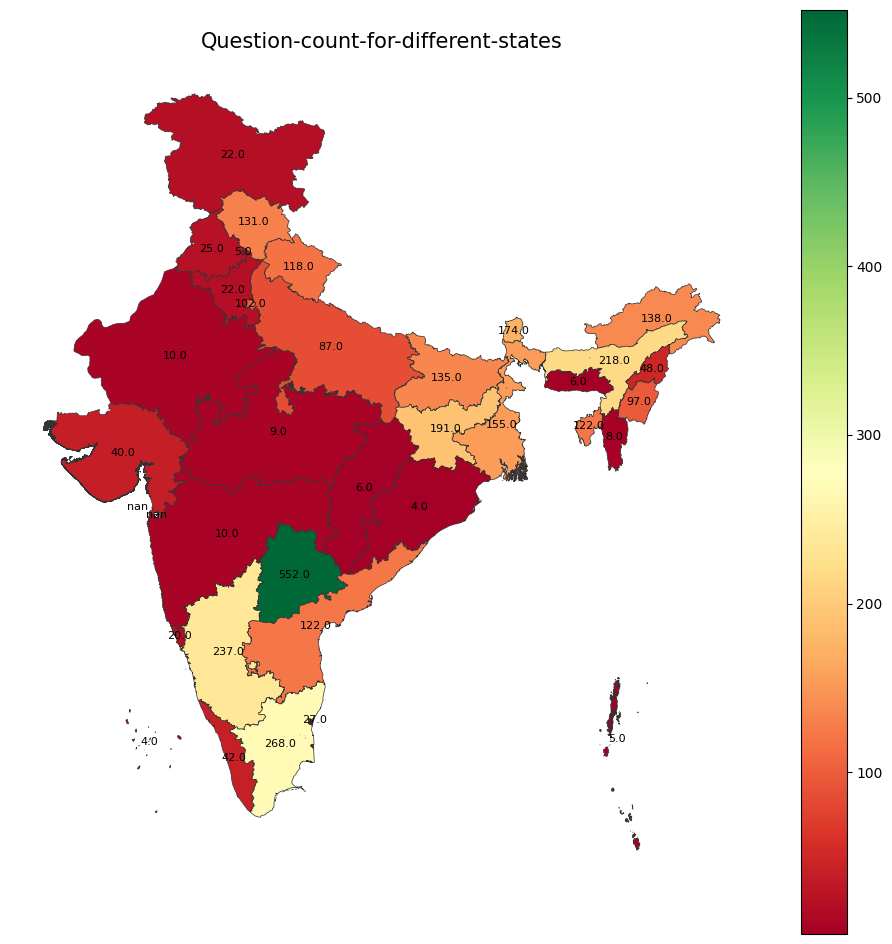}
  
  \caption{State-wise and attribute-wise distribution of questions in \textbf{\textit{VIRAASAT}}.}
  \label{fig:state-attr-dist}
\end{figure*}

\begin{table*}[t]
\centering
\scriptsize
\setlength{\tabcolsep}{5pt}
\renewcommand{\arraystretch}{1.15}
\begin{tabular}{
  >{\raggedright\arraybackslash}p{0.49\textwidth}
  >{\raggedright\arraybackslash}p{0.22\textwidth}
  >{\raggedright\arraybackslash}p{0.09\textwidth}
  >{\raggedright\arraybackslash}p{0.18\textwidth}
}
\toprule
\textbf{Question} &
\textbf{Graph Path (Anchor $\rightarrow$ State $\rightarrow$ Target)} &
\textbf{Answer} &
\textbf{Chain-of-Thought Trace} \\
\midrule

Which dish, a savoury fried bread made primarily from powdered rice, originated in the same state where the festival Baha Parab (the spring festival of the Santhal tribe) is celebrated?
&
\texttt{festival:baha\_parab} $\rightarrow$ \texttt{state:jharkhand} $\rightarrow$ \texttt{cuisine:dhooska}
&
Dhooska
&
\begin{minipage}[t]{\linewidth}
\vspace{-0.6em}
\begin{enumerate}
  \setlength{\itemsep}{0pt}
  \setlength{\parskip}{0pt}
  \setlength{\topsep}{0pt}
  \setlength{\partopsep}{0pt}
  \item Baha Parab is celebrated in Jharkhand.
  \item Dhooska originated in Jharkhand.
  \item Therefore, the dish is Dhooska.
\end{enumerate}
\vspace{0.2em}
\end{minipage}
\\
\midrule

Which dish, famous as a fermented sweet made with milk and sugar or jaggery, originated in the same state where the Garad saree is worn?
&
\texttt{costume:garad\_saree} $\rightarrow$ \texttt{state:west\_bengal} $\rightarrow$ \texttt{cuisine:mishti\_doi}
&
Mishti doi
&
\begin{minipage}[t]{\linewidth}
\vspace{-0.6em}
\begin{enumerate}
  \setlength{\itemsep}{0pt}
  \setlength{\parskip}{0pt}
  \setlength{\topsep}{0pt}
  \setlength{\partopsep}{0pt}
  \item Garad saree is from West Bengal.
  \item Mishti doi is from West Bengal.
  \item Therefore, the dish is Mishti doi.
\end{enumerate}
\vspace{0.2em}
\end{minipage}
\\
\midrule

Which sacred ghat, located on the banks of the Ganges and considered one of the most important pilgrimage sites, is situated in the same state where the ``Ramman festival'' is celebrated?
&
\texttt{festival:ramman\_festival} $\rightarrow$ \texttt{state:uttarakhand} $\rightarrow$ \texttt{tourism:har\_ki\_pauri}
&
Har ki Pauri
&
\begin{minipage}[t]{\linewidth}
\vspace{-0.6em}
\begin{enumerate}
  \setlength{\itemsep}{0pt}
  \setlength{\parskip}{0pt}
  \setlength{\topsep}{0pt}
  \setlength{\partopsep}{0pt}
  \item Ramman festival is celebrated in Uttarakhand.
  \item Har ki Pauri is in Uttarakhand.
  \item Therefore, the ghat is Har ki Pauri.
\end{enumerate}
\vspace{0.2em}
\end{minipage}
\\

\bottomrule
\end{tabular}
\caption{Sample data instances from \textbf{\textit{VIRAASAT}} dataset. Each question requires traversing a constrained 2-hop path from an anchor entity to a bridge state and then to the target artifact.}
\label{tab:viraasat-samples}
\end{table*}

\section{Related Works}
Recent works have developed a growing set of benchmarks for evaluating LLMs on Indian cultural knowledge and regional specificity. \textit{SANSKRITI} \cite{maji-etal-2025-sanskriti} provides large-scale text-only evaluation across Indian states and union territories via multiple-choice questions spanning diverse cultural attributes, and reports persistent gaps across regions and topics. \textit{DRISHTIKON} \cite{maji-etal-2025-drishtikon} builds upon this line of research by introducing a multimodal benchmark with broad geographic coverage and culturally grounded tasks that includes inference and alignment. Complementary efforts include multilingual evaluation suites such as \textit{L3Cube-IndicQuest} \cite{rohera-etal-2024-l3cube} and multimodal multilingual settings such as \textit{IndicVisionBench} dataset \cite{faraz2025indicvisionbenchbenchmarkingculturalmultilingual} that emphasize OCR, translation, and VQA under culturally grounded content. Other datasets target specific forms of cultural knowledge or difficulty; \textit{BhashaBench V1} \cite{devane2025bhashabenchv1comprehensivebenchmark} evaluates professional and traditional domains at scale, while \textit{ParamBench} \cite{maheshwari2025parambenchgraduatelevelbenchmarkevaluating} focuses on graduate-level Hindi questions designed to probe deeper cultural reasoning. Community-centered resources such as \textit{DOSA} \cite{seth-etal-2024-dosa} and conceptual inventories such as \textit{DIWALI} \cite{sahoo2025diwalidiversityinclusivityaware} further emphasize authenticity and sub-regional diversity. In contrast to these primarily single-turn factual or multiple-choice evaluations, VIRAASAT evaluates \emph{multi-hop} cultural reasoning in a multi-hop setting.

Beyond India-specific datasets, broader cultural benchmarks examine cross-cultural generalization and implicit value reasoning. Research on common sense evaluations spanning multiple countries and languages report systematic performance disparities that correlate with cultural and linguistic representation in training data \cite{Nguyen_2023}. Benchmarks explicitly targeting cultural reasoning, such as \textit{CQ-Bench} \cite{Liu2025CanLG} test implicit value inference, highlighting that deeper cultural inference remains challenging even for strong models. 



 \begin{table}[t]
\small
    \centering
    \resizebox{\linewidth}{!}{
    \begin{tabular}{lccc}
        \toprule
        \textbf{Dataset} & \textbf{\# Train samples} & \textbf{\# Val samples} & \textbf{\# Test samples}\\
        \midrule
        \textbf{\textit{VIRAASAT}} & 2252 & 323 & 643 \\ 
        \midrule
        \textbf{Total} & \multicolumn{3}{c}{\textbf{3218 questions}} \\
        \bottomrule
    \end{tabular}
    }
    \caption{Dataset statistics for \textbf{\textit{VIRAASAT}}.}
    \label{tab:dataset-stat}
\end{table}
\section{VIRAASAT Dataset Creation}

\begin{figure*}[t]
\centering
\includegraphics[width=0.9\textwidth]{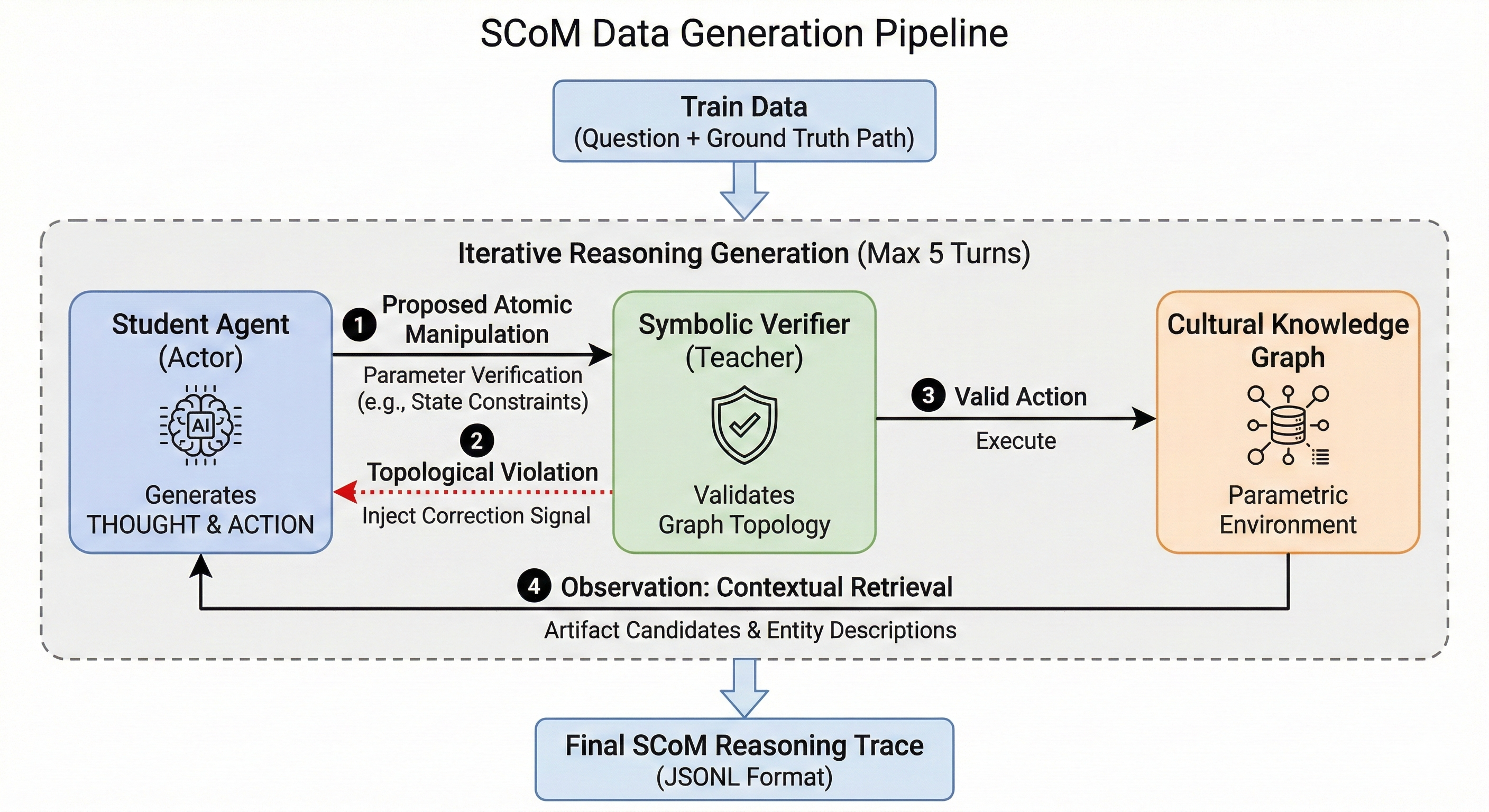} 
\caption{\textbf{SCoM reasoning-trace generation pipeline:} The figure illustrates the SCoM reasoning trace generation pipeline. The \emph{Student Agent} (actor) solves the question by producing an explicit, step-wise trace consisting of \texttt{THOUGHT} and \texttt{ACTION} steps, where each \texttt{ACTION} represents an \emph{atomic manipulation} over the cultural knowledge graph (entity grounding, enforcing the state constraint, candidate retrieval and resolution, and target resolution). Each proposed action is checked by the \emph{Symbolic Verifier} (teacher) that validates graph-topology consistency with the constrained path; if an action violates constraints, the verifier injects a correction signal that redirects the next step, preventing drift from the intended traversal. Valid actions are executed against the \emph{parametric environment} backed by the cultural knowledge graph, returning \texttt{OBSERVATION} outputs such as candidate entities and textual descriptors used to guide subsequent steps. The actor–verifier loop iterates for a bounded number of turns (k=5), producing a faithful, tool-grounded SCoM reasoning trace that is used for downstream SFT, teaching a model to perform structured graph-grounded retrieval and reasoning.}

\label{fig:scom-pipeline}
\end{figure*}

\subsection{Annotators}
The data construction process involved thirteen Indian university students belonging to different  states of India. The annotators possess an appropriate understanding of Indian culture along with English language proficiency. The annotators received rigorous training, including identifying appropriate culturally relevant artifacts,as well as how to write accurate descriptors and convert them into multi-hop reasoning questions.

\subsection{Annotator Training}
All the annotators participated in the following training protocol covering the entire workflow of our question curation pipeline.

\begin{enumerate}
    \item Gather and Extract Data from Indian cultural sources and government repositories.

    \item Format the data in order: State, Cultural Artifact, Descriptor for the artifact.

    \item Check the newly curated question for grammatical errors, reasoning steps,unwanted data and many more inconsistencies.
     
     \item Vote for the entry of the question into the dataset.
\end{enumerate}


\subsection{Data Collection}
The authors were tasked with collating relevant sources for data collection. These sources include URLs containing links to the description of the culturally specific elements, which are publicly available. Afterwards, the trained authors were asked to scrape the data within a 1-month period.
The task was to collect metadata, including the
Source link, state, artifact name, and descriptor for the corresponding cultural artifact.

\subsection{Knowledge Graph Construction}
We build a knowledge graph from the manually curated inventory of Indian cultural artifacts. Each artifact is represented as a unique node with its descriptor text stored as node metadata. We define a controlled set of relation types by grouping semantically similar attributes and mapping them to canonical predicates, enabling consistent connectivity across the graph. The resulting schema includes canonical relations from artifact to state and from artifact to attribute, yielding a graph with over 1600 edges.

\subsection{Question Generation}
We use the Knowledge Graph to generate template-based multi-hop questions with explicit symbolic grounding. Each 2-hop question is constructed around a state bridge: an anchor artifact $A$, its associated state $S$, and an answer artifact $B$, forming the path $A \rightarrow S \rightarrow B$. The correct reasoning trace used to construct each question is captured by a For example,
\texttt{graph\_path}: \url{festival:baha_parab|state:jharkhand|history:pathalgadi_movement}
 encodes the 2-hop chain from the anchor artifact to the answer artifact via the state bridge. The question includes two clues, one describing $A$ and one describing $B$. Clues are constrained to be informative yet not directly identifying, so that the answer cannot be inferred without using the bridge entity. For example, a clue such as ``the capital city of India'' is invalid because it uniquely determines the answer without requiring the state bridge.

We generate questions using a deterministic pipeline. First, we sample an anchor artifact $A$. We then retrieve its state $S$ and sample an answer artifact $B$ from the same state subject to a cross-attribute constraint (the attribute/theme of $B$ must differ from that of $A$). Clues are extracted from artifact descriptors using rule-based parsing (e.g., noun-phrase and pattern-based extraction) and inserted into a fixed template. To prevent dataset skew, we cap the maximum number of occurrences per artifact and enforce sampling constraints. This pipeline is automated and reproducible.

Finally, the artifact descriptors associated with the nodes on the \texttt{graph\_path}, together with the explicit path itself, are used to generate the ground-truth Chain-of-Thought trace for supervision using the \emph{gemini-2.0-flash model}.

\subsection{Manual Verification}
Template-based generation guarantees a valid structural path but does not ensure that each question is grammatically fluent or that its clues are appropriately non-specific. Rule-based clue extraction operates at the linguistic level and often fails to capture the context required to avoid accidental specificity or ambiguity. In our experiments, LLM-based approaches for clue selection and rewriting remain inconsistent for culturally nuanced entities and do not reliably enforce the non-specificity constraint. Therefore, manual annotators with cultural and linguistic expertise review and correct questions to ensure semantic validity and clarity.

\subsection{Question Curation}
After verification, annotators were required to provide a decision for every question, to determine its suitability for integration into the dataset. These decisions were based on the evaluation of both the correctness of the question and the authenticity of its cultural representation.


\subsection{Compensation}
We ensured competitive compensation for the annotators, exceeding the monthly average wage in each respective state.

\subsection{Factual Responsibility}
During data scraping, annotators were instructed to include questions accompanied by an answer and supporting facts and discard questions containing any kind of factual irregularity. 

\subsection{Inter-annotator Agreement}
After the creation of the initial dataset each and every question was cross-checked by multiple annotators and we used an anonymous voting mechanism to decide which questions would be kept in the dataset and hence we ensured an inter-annotator agreement of 0.92 (Cohen's Kappa).


\section{SCoM Reasoning Data Generation}
\label{sec:scom_trace_generation}

SCoM reasoning traces are synthesized to train a model to solve the given multi-hop question by explicitly \emph{simulating} a structured traversal over our underlying Knowledge Graph. SCoM replaces the standard CoT reasoning trace with a longer, agent-style trajectory that makes intermediate decisions observable and verifiable while remaining symbolically constrained. We cast reasoning as a controlled sequence of \textbf{Atomic Manipulations} over the graph: (i) grounding the anchor entity mentioned in the question, (ii) using the recovered bridge state to restrict the search space, and (iii) resolving the target by matching the question’s clue against candidate descriptors. Figure \ref{fig:scom-pipeline} highlights the key components of the SCoM pipeline.

\paragraph{\textbf{Actor agent:}}The actor agent (\emph{gemini-2.0-flash}) is prompted to behave like a Knowledge Graph agent that alternates between stating what information is missing and what operation to perform next, and requesting the corresponding retrieval step. The environment returns the result of that step as an observation (e.g., the anchor artifact's state or a state-conditioned candidate list), and the agent continues until it commits to the final answer. This representation exposes how the model narrows the hypothesis space and why the selected target is consistent with both the clue and the graph structure.

\paragraph{\textbf{Verifier Agent:}} A central requirement is that every intermediate step remains faithful to the intended graph traversal. To enforce this, we introduce a \textbf{mid-step oracle verifier} that monitors the agent’s actions against the bridge constraints implied by the question’s grounding path. The verifier performs step-level checks, ensuring that retrieval is conditioned on the correct bridge state and that search is conducted within the appropriate target category. When the actor deviates (e.g., queries the wrong state), the verifier injects a corrective interjection that explicitly identifies the inconsistency and forces the trace to continue under the corrected constraint. This design is intentional: end-of-trace verification only evaluates the final answer and permits arriving correctly for the wrong reasons, whereas mid-step interventions shape the trajectory itself, promoting disciplined, path-based reasoning and reducing spurious shortcuts. In effect, the verifier acts as a symbolic supervisor that repeatedly re-anchors generation to the graph topology at each stage.

\paragraph{\textbf{Knowledge Graph Environment:}} Observations are generated by an authentic simulator backed by the curated knowledge base, forcing the actor to perform sequential atomic manipulations on the candidate sets. The resulting SCoM dataset therefore, consists of long-form reasoning traces that are (i) structurally aligned with the atomic manipulation framework, (ii) grounded in graph-conditioned retrievals, and (iii) corrected online when the actor strays from the intended traversal. This yields supervision that trains the model to execute faithful multi-hop retrieval

\begin{figure}[t]
\centering
\begin{minipage}{0.98\linewidth}
\begin{tcolorbox}[
  title=\textbf{SCoM Trace Example},
  colback=blue!3,
  colframe=blue!75!black,
  colbacktitle=blue!75!black,
  coltitle=white,
  boxrule=0.6pt,
  arc=2pt,
  left=5pt,right=5pt,top=5pt,bottom=5pt,
  width=\linewidth
]
\scriptsize
\setlength{\tabcolsep}{4pt}
\renewcommand{\arraystretch}{1.15}
\begin{tabular}{p{0.16\linewidth} p{0.80\linewidth}}
\textbf{User} &
Which sacred ritual structure, used during the Nyokum festival where the Nyokum goddess is the principal deity, is celebrated in the same state where the art piece Khamti Pottery was crafted? \\

\textbf{THOUGHT} &
Find the origin state of \emph{Khamti Pottery}, then retrieve Nyokum-related festival artifacts from the same state and select the matching ritual structure. \\

\textbf{ACTION} &
\texttt{\detokenize{lookup_entity_by_name(name="Khamti Pottery")}} \\

\textbf{OBSERVATION} &
\texttt{\detokenize{{name: "Khamti Pottery", state: "Arunachal Pradesh"}}} \\

\textbf{REFLECTION} &
Restrict search to \emph{Arunachal Pradesh}. \\

\textbf{ACTION} &
\texttt{\detokenize{get_artifacts_in_state(state="Arunachal Pradesh",}}
\newline
\texttt{\detokenize{target_type="Festival")}} \\

\textbf{OBSERVATION} &
\texttt{\detokenize{... "Yugang Bamboo Altars" (Nyishi; Nyokum Goddess as principal deity) ...}} \\

\textbf{FINAL} &
\textbf{Yugang Bamboo Altars} \\

\textbf{Graph path} &
\texttt{\detokenize{art:khamti_pottery | state:arunachal_pradesh | festival:yugang_bamboo_altars}}
\end{tabular}
\end{tcolorbox}
\end{minipage}
\caption{Example SCoM trace (cleaned) illustrating tool-grounded retrieval over a constrained 2-hop path.}
\label{fig:scom-trace-example}
\end{figure}

\section{Experimental Setup}
\label{sec:exp_setup}

\subsection{Zero-shot Benchmarking}
\label{sec:inference_setup}
We evaluate the base model under a zero-shot prompting regime that requires explicit reasoning prior to producing the final prediction. The model receives only the task instruction and the question, and it generates a reasoning trace followed by a final answer in the prescribed format. Under this setting, the model is expected to infer the relevant state implied by the anchor entity and then identify the target artifact that is consistent with that state and the provided clue.

\subsection{Supervised Fine-tuning on CoT Data}
\label{sec:sft_setup}
To adapt the model to the dataset’s reasoning style and symbolic constraints, we perform Supervised Fine-Tuning (SFT) on the CoT training data using Parameter-Efficient Fine-Tuning (PEFT) \cite{houlsby2019parameterefficienttransferlearningnlp}. PEFT updates a small set of trainable adapter parameters while keeping the base model weights fixed. Training is performed for 5 epochs on the train split and the final results are reported on the test split. The key hyperparameter settings are shown in Table \ref{tab:hyperparameters}.


\begin{figure}[t]
\centering
\begin{tcolorbox}[
  width=\columnwidth,
  colback=orange!6!white,          
  colframe=orange!75!black,        
  colbacktitle=orange!75!black,    
  coltitle=white,
  title=\textbf{SFT Prompt},
  fonttitle=\footnotesize\bfseries,
  fontupper=\footnotesize,
  boxrule=0.6pt,
  arc=1.5pt,
  left=4pt,right=4pt,top=3pt,bottom=3pt
]
\textbf{Role:} You are an expert in Indian cultural heritage, traditions, and general knowledge. Your task is to identify the specific cultural entity or concept described by the user.

\vspace{2pt}
The entity falls under one of these categories:
\vspace{2pt}

\begin{itemize}
\setlength{\itemsep}{0pt}
\setlength{\topsep}{2pt}
  \item Tourism \& History (Sites, Monuments, Forts, Caves)
  \item Religion \& Rituals (Temples, Ceremonies, Sacred Sites)
  \item Art, Dance \& Music (Paintings, Artifacts, Folk Forms)
  \item Festivals \& Cultural Common Sense
  \item Personalities (Historical Figures, Artists)
  \item Cuisine, Costume \& Medicine
  \item Sports, Transport \& Language
\end{itemize}

\vspace{4pt}
Output rules:
\begin{enumerate}
\setlength{\itemsep}{0pt}
\setlength{\topsep}{2pt}
  \item First write reasoning inside \texttt{<think>}...\texttt{</think>}
  \item Then write the final answer inside \texttt{<answer>}...\texttt{</answer>}
  \item Do not write anything else
\end{enumerate}
\end{tcolorbox}

\caption{Prompt used for SFT, providing the necessary task context and details.}
\label{fig:sft-cot-prompt}
\end{figure}

\subsection{Supervised Fine-tuning on SCoM Data}
\label{sec:sft_scom_setup}
We additionally fine-tune the model on the SCoM reasoning-trace data, which provides long-form, step-structured supervision for path-faithful reasoning. In this setting, the training target includes the full SCoM trace and the final answer, encouraging the model to internalize the sequence of atomic manipulations required to traverse the knowledge graph under the bridge-state constraint. We perform SFT using PEFT, updating only lightweight adapter parameters while keeping the base model frozen. Training is performed for 5 epochs on the train split and final results are reported on the \textbf{test} split.

\subsection{Evaluation Metrics}
\label{sec:metrics}
We report three exact-match style metrics that capture correctness of both the intermediate bridge entity and the final answer:
\begin{itemize}
    \item \textbf{State Match (\%):} Accuracy of the predicted \emph{bridge entity}, defined as the state used to connect the anchor artifact to the target artifact. 
    \item \textbf{Answer Match (\%):} Accuracy of the predicted \emph{final artifact} (the answer entity). 
    \item \textbf{Full Match (\%):} Accuracy when \emph{both} the bridge state and the final artifact are both correct for the same example. 
\end{itemize}

\begin{table*}[t]
\centering
\footnotesize
\setlength{\tabcolsep}{5pt}
\renewcommand{\arraystretch}{1.05}

\resizebox{0.75\textwidth}{!}{%
\begin{tabular}{l r r r}
\toprule
\textbf{Model} & \textbf{State Match (\%)} & \textbf{Answer Match (\%)} & \textbf{Full Match (\%)} \\
\midrule
\multicolumn{4}{l}{\textbf{Zero-shot}} \\
\midrule
gemini-2.0-flash & 81.71 & 35.19 & 31.32 \\
GPT4.1-mini & 77.36  & 34.57 & 32.56 \\
Qwen2.5-3B-Instruct & 25.43 & 3.72 & 2.17 \\
Qwen3-4B-Thinking-2507 & 7.44 & 1.71 & 0.31 \\
Qwen2.5-7B-Instruct & 43.88 & 3.72 & 1.71 \\
Phi-4-mini-reasoning & 35.97 & 8.84 & 6.05 \\
OpenHathi-7B-Hi-v0.1-Base & 25.27 & 4.19 & 2.17 \\
sarvam-1 & 11.78 & 7.13 & 2.17 \\
Llama-3.1-8B-Instruct & 55.81 & 11.78 & 10.08 \\
Mistral-7B-Instruct-v0.3 & 57.05 & 15.04 & 12.25 \\
gemma-2b-it & 24.34 & 4.96 & 3.26 \\
\midrule
\multicolumn{4}{l}{\textbf{SFT (CoT)}} \\
\midrule
Qwen2.5-3B-Instruct & 87.87 & 34.73 & 34.26 \\
Qwen2.5-7B-Instruct & 89.11 & 37.33 & 36.70 \\
Qwen3-4B-Thinking-2507 & 88.22 & 37.36 & 35.50 \\
Phi-4-mini-reasoning & 84.19 & 30.08 & 28.53 \\
\midrule
\multicolumn{4}{l}{\textbf{SFT (SCoM)}} \\
\midrule
Qwen2.5-3B-Instruct & 87.25 & 51.48 & 50.70 \\
\rowcolor{blue!10}
\textcolor{blue}{\textbf{Qwen2.5-7B-Instruct}} &
\textcolor{blue}{\textbf{91.45}} &
\textcolor{blue}{\textbf{58.01}} &
\textcolor{blue}{\textbf{57.54}} \\
Phi-4-mini-reasoning & 81.03 & 43.55 & 43.39 \\
\bottomrule
\end{tabular}%
}

\caption{Benchmarking results of various models on \textbf{\textit{VIRAASAT}} dataset across different settings (Zero-shot inference, SFT on CoT data, SFT on SCoM data) on the test set. Reported metrics are State Match \%, Answer Match \% and Full Match \% respectively. The best results per column are highlighted in bold.}
\label{tab:t1-viraasat-benchmark}
\end{table*}

\begin{table}[t]
  \centering
  \caption{Training Hyperparameters (SFT CoT)}
  \label{tab:hyperparameters}
  \begin{tabular}{|l|c|}
    \hline
    \textbf{Hyperparameter} & \textbf{Value} \\
    \hline
    \texttt{max\_new\_tokens} & 512 \\
    \texttt{temperature} & 0.0 \\
    \texttt{learning\_rate} & $5\times10^{-5}$ \\
    \texttt{optimizer} & \texttt{paged\_adamw\_8bit} \\
    \texttt{gradient\_accumulation\_steps} & 4 \\
    \texttt{lora\_alpha} & 32 \\
    \texttt{lora\_r} & 16 \\
    \hline
  \end{tabular}
\end{table}

\section{Results and Discussion}

\subsection{Zero-shot Results}
Across model categories, zero-shot performance on \textbf{\textit{VIRAASAT}} remains limited, indicating that multi-hop cultural reasoning over long-tail entities is not reliable. Among closed-source LLMs, \textit{gemini-2.0-flash} attains the strongest zero-shot results (State Match 81.71\%, Answer Match 35.19\%, Full Match 31.32\%), suggesting stronger generalization under culturally grounded constraints. In contrast, open-weight LLMs exhibit substantially lower end-to-end accuracy: \textit{Qwen2.5-7B-Instruct} reaches 43.88\% State Match but only 3.72\% Answer Match, implying that models often identify the correct bridge state yet fail to resolve the final target artifact. This recurrent gap between State Match and Full Match highlights that \textbf{\textit{VIRAASAT}} requires both discovering the correct intermediate constraint (state) and completing the target retrieval and semantic resolution.

Reasoning-oriented models show mixed behavior in the zero-shot regime. \textit{Phi-4-mini-reasoning} achieves comparatively stronger Answer Match (8.84\%) and Full Match (6.05\%) than similarly sized baselines, while \textit{Qwen3-4B-Thinking-2507} remains weak across metrics, indicating that nominal reasoning capability does not necessarily translate to faithful multi-hop cultural traversal. Overall, Indic models and SLMs remain the weakest in zero-shot settings, consistent with limited pretraining exposure to the long-tail cultural entities and region-specific artifacts emphasized by \textbf{\textit{VIRAASAT}}.

\subsection{SFT Results}
Supervised fine-tuning (SFT) yields large improvements over zero-shot inference, particularly in intermediate state prediction. Under SFT on CoT traces, State Match rises into the high-80s for several open-weight backbones (e.g., \textit{Qwen2.5-7B-Instruct}: 89.11\%, \textit{Qwen2.5-3B-Instruct}: 87.87\%, \textit{Qwen3-4B-Thinking-2507}: 88.22\%). However, gains in Answer Match and Full Match remain moderate and cluster in the mid-30s (Answer Match up to 37.36\%, Full Match up to 36.70\%). This indicates that CoT supervision primarily helps models learn the structural regularities needed to predict the bridge constraint, but does not consistently enforce faithful retrieval and verification of the target artifact.

\noindent\textbf{SCoM fine-tuning.}
Training on SCoM traces produces the strongest and most consistent improvements, especially on the end-to-end metrics that require resolving the final target. \textit{Qwen2.5-7B-Instruct} achieves the best overall performance (State Match 91.45\%, Answer Match 58.01\%, Full Match 57.54\%), improving Full Match by +20.84 points over its CoT-SFT counterpart (36.70\% $\rightarrow$ 57.54\%). \textit{Qwen2.5-3B-Instruct (SCoM)} similarly improves (Full Match 50.70\% vs.\ 34.26\% with CoT-SFT), and \textit{Phi-4-mini-reasoning (SCoM)} reaches 43.39\% Full Match. Notably, SCoM narrows the gap between intermediate correctness and final correctness: increases in Answer Match track increases in State Match more closely than in CoT-SFT, suggesting that the model learns a more faithful execution pattern. These results support the premise that symbolically grounded, verifier-aligned manipulation traces are more effective than generic CoT supervision for multi-hop cultural reasoning.

\paragraph{Model-category trends.}
Closed-source LLMs lead in the zero-shot setting, reflecting stronger generalization under distribution shift. Open-weight LLMs and LRMs benefit substantially from supervised adaptation, with the largest end-to-end gains arising from SCoM supervision Vis-à-vis CoT, indicating that enforcing symbolic grounding during training improves target resolution under constraints. Indic models and SLMs remain comparatively weak in zero-shot settings, consistent with the long-tail nature of the entities; nevertheless, the SFT results show that improved supervision can partially mitigate pretraining limitations, with the strongest outcomes achieved by capable open-weight backbones trained on SCoM traces.

Finally, the results collectively indicate that the primary bottleneck on our dataset is not predicting the intermediate state but reliably resolving the target artifact under the correct constraint. CoT-SFT improves bridge prediction substantially but only partially addresses target resolution, whereas SCoM-SFT directly targets this failure mode by training models to execute graph-grounded manipulations, yielding large gains in Answer Match and Full Match.

\textbf{OOD Evaluation:} We evaluate OOD transfer by testing SCoM-tuned Qwen2.5 models on a 10 \% SANSKRITI subset (2,183 samples) with constrained multiple-choice decoding. Table \ref{tab:sanskriti-ood} shows that SCoM matches the vanilla baselines, with a small gain for Qwen2.5-7B and comparable performance for 3B. This indicates SCoM preserves factual-cultural recall on MCQ-style benchmarks while primarily benefiting open-ended multi-hop reasoning.

\begin{table}[t]
\centering
\scriptsize
\setlength{\tabcolsep}{5pt}
\renewcommand{\arraystretch}{1.05}

\begin{tabular}{l r r r}
\toprule
\textbf{Model (Qwen2.5-7B-Instruct)} & \textbf{State match (\%)} & \textbf{Answer match (\%)} & \textbf{Full Match (\%)} \\
\midrule
\rowcolor{blue!10}
\textcolor{blue}{\textbf{SCoM (all components)}} &
\textcolor{blue}{\textbf{88}} &
\textcolor{blue}{\textbf{52}} &
\textcolor{blue}{\textbf{52}} \\
 w/o critic & 75 & 37 & 36 \\
 w/o manipulations & 67 & 12 & 9 \\
\bottomrule
\end{tabular}

\caption{Ablation results for SCoM showing the impact of critic agent and atomic manipulations respectively. Trained on 1k samples and evaluated on 100 test samples.}
\label{tab:scom-ablation}
\end{table}

\begin{table}[t]
    \centering
    \scriptsize
    \setlength{\tabcolsep}{4pt}
    \renewcommand{\arraystretch}{1.15}
    \begin{tabular}{l c}
    \toprule
    \textbf{Model} & \textbf{Attribute-level Accuracy (\%)} \\
    \midrule
    Qwen2.5-3B-Instruct & 75.57 \\
    Qwen2.5-7B-Instruct & 80.84 \\
    Qwen2.5-3B-Instruct (SCoM) & 79.25 \\
    \rowcolor{blue!10}
    \textcolor{blue}{\textbf{Qwen2.5-7B-Instruct (SCoM)}} &
    \textcolor{blue}{\textbf{84.97}} \\
    \bottomrule
    \end{tabular}
    \caption{OOD Evaluation Results on 10 \% SANSKRITI subset. Attribute-level accuracy \% is reported.}
    \label{tab:sanskriti-ood}
\end{table}

\section{Ablations}
\label{sec:ablations}
To isolate the contributions of specific components within our framework, we perform two ablations on \modelname{}\allowbreak under \textsc{SCoM} supervision (1k train; 100 test). These experiments test whether gains primarily arise from (i) verified step-level supervision or (ii) the explicit, manipulation-centric reasoning format. The results are presented in Table \ref{tab:scom-ablation}.

\paragraph{\textbf{Ablation 1: Removal of Symbolic Verifier Agent:}}
\label{sec:ablation_unverified}
This setting evaluates the necessity of mid-step symbolic verification in the student-teacher synthesis pipeline. We generate an alternative training set by disabling the symbolic verifier (critic), such that the actor produces traces without external feedback. Table~\ref{tab:scom-ablation} shows that removing verification reduces State Match from 88\% to 75\% and Answer Match from 52\% to 37\%, leading to a 16-point drop in Full Match (52\% $\rightarrow$ 36\%). This indicates that step-level verification is important to prevent reasoning drift and maintain path-faithful supervision during fine-tuning.

\paragraph{\textbf{Ablation 2: Removal of Atomic Manipulations:}}
\label{sec:ablation_no_tools}
This setting evaluates the impact of explicit atomic manipulations (tool simulations) on grounding and controllability. Starting from the verified SCoM dataset, we remove all \texttt{ACTION} (tool call) and \texttt{OBSERVATION} (tool output) spans, retaining only \texttt{THOUGHT} and \texttt{REFLECTION}. This converts SCoM traces into implicit CoT-style rationales without state-conditioned retrieval steps. Table~\ref{tab:scom-ablation} shows a sharp degradation: State Match drops from 88\% to 67\%, Answer Match drops from 52\% to 12\%, and Full Match collapses from 52\% to 9\%. This suggests that explicit manipulations are the primary driver of correct target resolution under the constrained 2-hop setting.

\section{Error Analysis}
We analyze a representative set of failure cases to examine where models fail on \textbf{\textit{VIRAASAT}}. Figures~\ref{fig:sft(cot)-error-example1} and~\ref{fig:scom-success-example1} illustrate a recurring pattern wherein models often recover the correct bridge state entity, but fail to reliably execute the second reasoning step needed to identify the target artifact. In these cases, the model typically substitutes the correct answer with a closely related or non-existent artifact. In contrast, SCoM-trained models reduce this error mode by explicitly decomposing the problem into grounded atomic manipulations and retrieving the correct artifacts.

\begin{figure}[t]
\centering
\begin{minipage}{0.98\linewidth}
\begin{tcolorbox}[
  title=\textbf{Error Example (Qwen2.5-7B (SFT(CoT)))},
  colback=red!3,
  colframe=red!75!black,
  colbacktitle=red!75!black,
  coltitle=white,
  boxrule=0.6pt,
  arc=2pt,
  left=5pt,right=5pt,top=5pt,bottom=5pt,
  width=\linewidth
]
\scriptsize
\setlength{\tabcolsep}{4pt}
\renewcommand{\arraystretch}{1.15}
\begin{tabular}{p{0.16\linewidth} p{0.80\linewidth}}
\textbf{User} &
 Which dish, famous for mithai prepared with only three ingredients, originated from the same state where the festival Mysore Dasara is celebrated?
\\

\textbf{THOUGHT} &
Mysore Dasara is a festival celebrated in the state of Karnataka. Pocharellu is a mithai made with only three ingredients that originates from Karnataka. Therefore, Pocharellu is the dish that originated in the same state where Mysore Dasara is celebrated and is known for being a mithai made with only three ingredients. \\

\textbf{GT Answer} &
\texttt{\detokenize{{name: "Mysore Pak", state: "Karnataka"}}} \\

\textbf{Model-Output} &
\texttt{\detokenize{{name: "Pocharellu", state: "Karnataka"}}} \\

\textbf{Sate-Identification} &
\texttt{\detokenize{The Model is able to identify the state using the Anchor Artifact which is Mysuru Dasara a festival widely celebrated in the Indian State of Karnataka.}} \\

\textbf{Artifact-Identification} &
\texttt{\detokenize{Unfortunately after the identification of the state the model is unable to further reason through the identifier which is causing the model to hallucinate.}} \\

\end{tabular}
\end{tcolorbox}
\end{minipage}
\caption{Sample error analysis for SFT (CoT). The model infers the correct state entity but hallucinates the final answer artifact.}
\label{fig:sft(cot)-error-example1}
\end{figure}

\begin{figure}[t]
\centering
\begin{minipage}{0.98\linewidth}
\begin{tcolorbox}[
  title=\textbf{Correct Example (Qwen2.5-7B (SCoM))(Ours)},
  colback=green!3,
  colframe=green!75!black,
  colbacktitle=green!75!black,
  coltitle=white,
  boxrule=0.6pt,
  arc=2pt,
  left=5pt,right=5pt,top=5pt,bottom=5pt,
  width=\linewidth
]
\scriptsize
\setlength{\tabcolsep}{4pt}
\renewcommand{\arraystretch}{1.15}
\begin{tabular}{p{0.16\linewidth} p{0.80\linewidth}}
\textbf{User} &
 Which dish, famous for mithai prepared with only three ingredients, originated from the same state where the festival Mysore Dasara is celebrated?
\\

\textbf{GT Answer} &
\texttt{\detokenize{{name: "Mysore Pak", state: "Karnataka"}}} \\

\textbf{THOUGHT} &
The question asks for a dish that uses only three ingredients and originates from the same state as the Mysore Dasara festival. I need to first find the state where Mysore Dasara is celebrated. Then, I can search for dishes originating from that state that use only three ingredients. I will start by looking up the entity "Mysore Dasara". \\

\textbf{ACTION} &
\texttt{\detokenize{lookup_entity_by_name(name="Mysore Dasara")}} \\

\textbf{OBSERVATION} &
\texttt{\detokenize{{'name': 'Mysore Dasara', 'state': 'Karnataka'}}} \\

\textbf{REFLECTION} &
Restrict search to \emph{Karnataka}. \\

\textbf{ACTION} &
\texttt{\detokenize{get_artifacts_in_state(state="Karnataka",}}
\newline
\texttt{\detokenize{target_type="Cuisine")}} \\

\textbf{OBSERVATION} &
\texttt{\detokenize{... "Mysore Pak" (A type of mithai prepared with only three ingredients: ghee, sugar and besan.), "Maddur Vada specialty" (A savoury fritter-type snack made of rice flour, semolina and maida flour mixed with onion, curry leaves, green chillies, grated coconut, cashews / groundnuts, ghee, salt, and asafoetida. The onion and curry leaves are fried and then mixed with water to make a soft dough, which is made into patties and deep fried)...}} \\
\textbf{THOUGHT} &
From the list of cuisines in Karnataka, "Mysore Pak" is described as a type of mithai prepared with only three ingredients: ghee, sugar, and besan. This matches the description in the question. \\

\textbf{FINAL} &
\textbf{Mysore Pak} \\

\textbf{Model-Output} &
\texttt{\detokenize{{name: "Mysore Pak", state: "Karnataka"}}} \\

\end{tabular}
\end{tcolorbox}
\end{minipage}
\caption{Sample error analysis for SFT (SCoM). The model accurately answers the bridge state and final answer artifact.}
\label{fig:scom-success-example1}
\end{figure}

 \section{Limitations and Future Work}
Our study has several limitations that motivate clear extensions. First, \textbf{\textit{VIRAASAT}} is built around a fixed 2-hop template where the state/UT acts as the bridge entity. While this yields controlled multi-hop evaluation, it does not cover richer graph patterns (e.g. 3-hop chains or alternate bridge entity types). Second, the underlying knowledge graph is intentionally lightweight and centered on the selected cultural attributes, which makes it comparatively sparse and may omit many culturally salient relations that appear in denser heritage KGs. Third, despite careful curation, some artifacts naturally span multiple states or have shared regional variants, which can introduce ambiguity when context in the question is insufficient to uniquely pin down a single state association. Finally, the benchmark is currently English-only; it does not evaluate state-specific multilingual queries or code-mixed usage that is common in Indian contexts.

Future work will expand \textbf{\textit{VIRAASAT}} along three directions: (i) broader graph coverage and greater question diversity, including additional cultural attributes, denser relations, multi-bridge templates, and longer-hop questions; (ii) the addition of multilingual and state-localized question variants to examine multilingual multi-hop reasoning; and (iii) extending evaluations beyond SFT by benchmarking retrieval-augmented generation (RAG) and Reinforcement Learning based algorithms (e.g., RLVR-style optimization like GRPO) that reward path-faithful reasoning.

\section{Conclusion}
In this work, we introduce \textbf{\textit{VIRAASAT}}, a textual multi-hop question-answering benchmark for Indian culture, comprising over 3,200 questions spanning diverse cultural artifacts across 28 states and 8 Union Territories. Each question is paired with verifiable supervision derived from our curated cultural knowledge graph, enabling reliable and scalable evaluation of whether models can connect multiple grounded cultural facts to reach the correct answer. Our experiments show that zero-shot prompting remains unreliable for this task, and that supervised fine-tuning on standard chain-of-thought traces improves performance but does not consistently yield correct end-to-end reasoning. To bridge this gap, we propose \textbf{Symbolic Chain-of-Manipulation (SCoM)}, a structured supervision paradigm that decomposes multi-hop reasoning into a sequence of \emph{symbolically grounded atomic manipulations} over the knowledge graph. SCoM trains models on verifier-corrected reasoning traces in which each intermediate step corresponds to a valid graph-constrained operation, yielding controllable and path-faithful supervision.
 Across models, SCoM consistently improves all evaluation metrics and achieves over 20\% gains relative to CoT fine-tuning. By releasing \textbf{\textit{VIRAASAT}} and our SCoM methodology, we provide a scalable benchmark and an effective training signal for advancing culturally grounded multi-hop reasoning in Indian contexts, supporting the development of better culturally aware and inclusive language technologies.





\appendix
\section{Additional Prompts}
\label{sec:appendix-prompts}

\begin{figure}[t]
\centering
\begin{tcolorbox}[
  width=\columnwidth,
  colback=purple!6!white,          
  colframe=purple!70!black,        
  colbacktitle=purple!70!black,    
  coltitle=white,
  title=\textbf{SCoM Actor Agent Prompt},
  fonttitle=\footnotesize\bfseries,
  fontupper=\footnotesize,
  boxrule=0.6pt,
  arc=1.5pt,
  left=4pt,right=4pt,top=3pt,bottom=3pt
]
\textbf{Role:} You are an expert \emph{Knowledge Graph Agent} solving questions about Indian culture.

\vspace{2pt}
\textbf{Available tools (simulate their usage):}
\begin{itemize}
\setlength{\itemsep}{0pt}
\setlength{\topsep}{2pt}
  \item \texttt{lookup\_entity\_by\_name(name)}: find an entity by exact name; returns state and description.
  \item \texttt{resolve\_entity\_by\_description(description)}: ground an entity from its descriptor text.
  \item \texttt{get\_artifacts\_in\_state(state, target\_type)}: retrieve state-conditioned candidates by category.
\end{itemize}

\vspace{2pt}
\textbf{Strict step format (repeat until termination):}
\begin{itemize}
\setlength{\itemsep}{0pt}
\setlength{\topsep}{2pt}
  \item \texttt{THOUGHT:} what you know, what is missing, and what you will do next.
  \item \texttt{ACTION:} one tool call (e.g., \texttt{lookup\_entity\_by\_name(name="Taj Mahal")}).
\end{itemize}

\vspace{2pt}
\textbf{Constraints:}
\begin{itemize}
\setlength{\itemsep}{0pt}
\setlength{\topsep}{2pt}
  \item Stop after \texttt{ACTION} and wait for \texttt{OBSERVATION}.
  \item If you receive \texttt{ORACLE INTERVENTION}, acknowledge the error and correct immediately.
  \item Continue until \texttt{FINAL ANSWER: <your answer>}.
\end{itemize}

\vspace{2pt}
\textbf{Reasoning rules:}
\begin{enumerate}
\setlength{\itemsep}{0pt}
\setlength{\topsep}{2pt}
  \item Identify the anchor entity mentioned in the question.
  \item Use the recovered state to constrain retrieval of candidates.
  \item If a search fails, reflect and try an alternative grounding or retrieval step.
  \item Use precise entity names and state names.
\end{enumerate}
\end{tcolorbox}

\caption{Prompt used for the SCoM Actor agent to generate structured reasoning traces under a strict \texttt{THOUGHT/ACTION/OBSERVATION} protocol.}
\label{fig:scom-actor-prompt}
\end{figure}

\bibliographystyle{ACM-Reference-Format}
\bibliography{references}
\end{document}